\documentclass[10pt,twocolumn,letterpaper]{article}

\usepackage{cvpr}              
\usepackage[accsupp]{axessibility}

\usepackage{bbm}
\usepackage{graphicx}
\usepackage{amsmath}
\usepackage{amssymb}
\usepackage{booktabs}
\usepackage{multirow}
\usepackage[ruled,linesnumbered]{algorithm2e}

\usepackage[pagebackref,breaklinks,colorlinks]{hyperref}

\usepackage[capitalize]{cleveref}
\crefname{section}{Sec.}{Secs.}
\Crefname{section}{Section}{Sections}
\Crefname{table}{Table}{Tables}
\crefname{table}{Tab.}{Tabs.}

\def\cvprPaperID{2616} 
\def\confName{CVPR}
\def\confYear{2023}

\begin{document}

\title{Center Focusing Network for Real-Time LiDAR Panoptic Segmentation}

\author{Xiaoyan Li$^{1,2}$ \hspace{5pt} Gang Zhang$^{3}$ \hspace{5pt} Boyue Wang$^{1,2}$ \hspace{5pt} Yongli Hu$^{1,2}$ \hspace{5pt} Baocai Yin$^{1,2}$\\
$^1$Beijing Municipal Key Lab of Multimedia and Intelligent Software Technology\\
$^2$Beijing Institute of Artificial Intelligence, Faculty of Information Technology,\\
Beijing University of Technology, Beijing 100124, China\\
$^3$Mogo Auto Intelligence and Telemetics Information Technology Co. Ltd.\\
{\tt\small \{xiaoyan.li,wby,huyongli,ybc\}@bjut.edu.cn \hspace{5pt} zhanggang11021136@gmail.com}
}

\maketitle

\begin{abstract}
LiDAR panoptic segmentation facilitates an autonomous vehicle to comprehensively understand the surrounding objects and scenes and is required to run in real time. The recent proposal-free methods accelerate the algorithm, but their effectiveness and efficiency are still limited owing to the difficulty of modeling non-existent instance centers and the costly center-based clustering modules. To achieve accurate and real-time LiDAR panoptic segmentation, a novel center focusing network (CFNet) is introduced. Specifically, the center focusing feature encoding (CFFE) is proposed to explicitly understand the relationships between the original LiDAR points and virtual instance centers by shifting the LiDAR points and filling in the center points. Moreover, to leverage the redundantly detected centers, a fast center deduplication module (CDM) is proposed to select only one center for each instance. Experiments on the SemanticKITTI and nuScenes panoptic segmentation benchmarks demonstrate that our CFNet outperforms all existing methods by a large margin and is 1.6 times faster than the most efficient method. The code is available at \url{https://github.com/GangZhang842/CFNet}.
\end{abstract}

\section{Introduction}
\label{sec:intro}
Panoptic segmentation~\cite{kirillov2019panoptic} combines both semantic segmentation and instance segmentation in a single framework. It predicts semantic labels for the uncountable \emph{stuff} classes (\eg \emph{road}, \emph{sidewalk}), while it simultaneously provides semantic labels and instance IDs for the countable \emph{things} classes (\eg \emph{car}, \emph{pedestrian}). The LiDAR panoptic segmentation is one of the bases for the safety of autonomous driving, which employs the point clouds collected by the Light Detection and Ranging (LiDAR) sensors to effectively depict the surroundings. Existing LiDAR panoptic segmentation methods first conduct semantic segmentation, and then achieve instance segmentation for the \emph{things} categories in two ways, the proposal-based and proposal-free methods.

\begin{figure}[!t]
\centering
\includegraphics[width=0.9\linewidth]{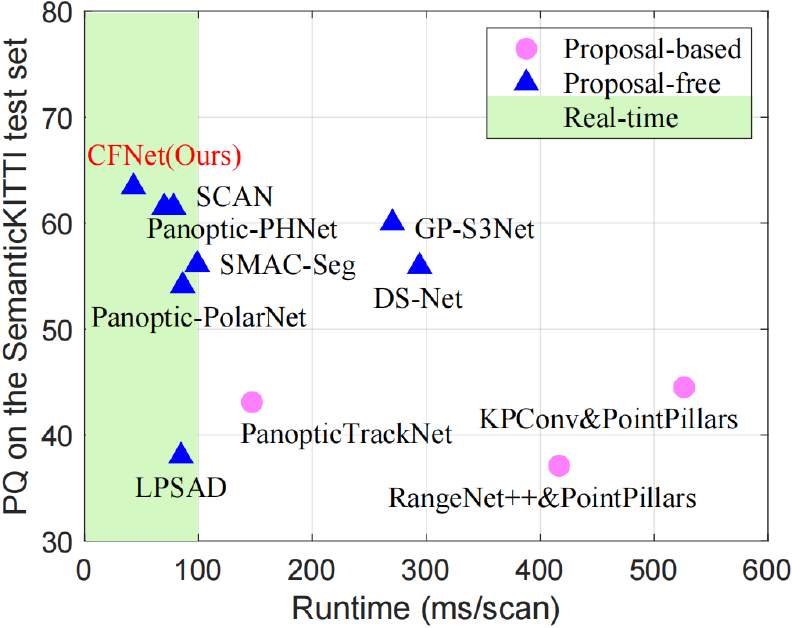}
\caption{PQ vs. runtime on the SemanticKITTI test set. Runtime measurements are taken on a single NVIDIA RTX 3090 GPU. The panoptic quality (PQ) is introduced in section~\ref{subsec_exp_setup}.}
\label{fig_prob_pq_speed}
\end{figure}

The proposal-based methods~\cite{sirohi2021efficientlps, hurtado2020mopt, ye2022lidarmultinet} adopt a two-stage process similar to the well-known Mask R-CNN~\cite{he2017mask} in the image domain. It first generates object proposals for the \emph{things} points by using 3D detection networks~\cite{lang2019pointpillars, shi2020pv} and then refines the instance segmentation results within each proposal. As shown in Fig.~\ref{fig_prob_pq_speed}, these methods are usually complicated and hardly achieve real-time processing, owing to their sequential multi-stage pipelines.

\begin{figure}[!t]
\centering
\includegraphics[width=0.85\linewidth]{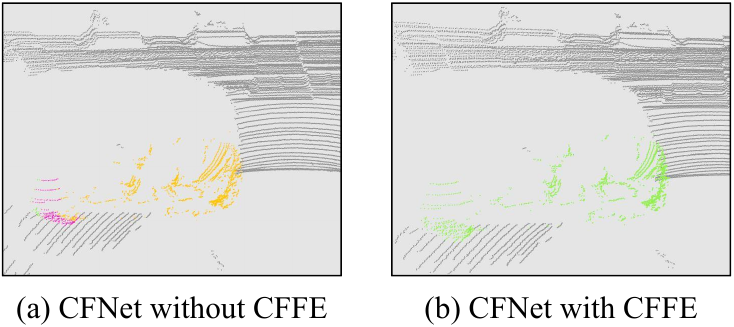}
\caption{Instance segmentation of a car. Without our CFFE, the car is split into parts (a), while the CFFE significantly alleviates this problem (b). Different colors represent different instances.}
\label{fig_cffe_effect}
\end{figure}

The proposal-free frameworks~\cite{hong2021lidar, zhou2021panoptic, li2021cpseg, gasperini2021panoster, razani2021gp, xu2022sparse, li2022panoptic} are more compact. To associate the \emph{things} points with instance IDs, these methods usually leverage the instance centers. Specifically, they regress the offsets from the points to their corresponding centers, and then adopt the class-agnostic center-based clustering modules~\cite{hong2021lidar, gasperini2021panoster, razani2021gp} or the bird’s-eye view (BEV) center heatmap~\cite{zhou2021panoptic, xu2022sparse, li2022panoptic}. However, two problems exist in these methods. First, for center feature extracting and center modeling, the non-existent instance centers increase the difficulty, considering that the LiDAR points are usually surface-aggregated~\cite{xu2022sparse} and an instance center is imaginary in most cases. As shown in Fig.~\ref{fig_cffe_effect}(a), the difficulty often results in the fault that one instance is incorrectly split into several parts. Second, for exploiting the redundantly detected centers, the clustering modules (\eg MeanShift, DBSCAN) are too time-consuming to support the real-time autonomous driving perception systems, while the BEV center heatmap cannot distinguish objects with different altitudes in the same BEV grid.

For accurate and fast LiDAR panoptic segmentation, a proposal-free center focusing network (CFNet) is proposed. For better encoding center features, a novel center focusing feature encoding (CFFE) is proposed to generate center-focusing feature maps by shifting the \emph{things} points to fill in the non-existent instance centers for more accurate predictions (as shown in Fig.~\ref{fig_cffe_effect}(b)). For center modeling, the CFNet not only decomposes the panoptic segmentation task into the widely-used semantic segmentation and center offset regression, but also proposes a new confidence score prediction for indicating the accuracy of the center offset regression. Subsequently, for the detected centers exploiting, a novel center deduplication module (CDM) is designed to select one center for a single instance. The CDM keeps the predicted centers with higher confidence scores, while suppressing the ones with lower confidence. Finally, instance segmentation is achieved by assigning the shifted \emph{things} points to the closest center.
For efficiency, the proposed CFNet is built on the 2D projection-based segmentation paradigm. Our contributions are as follows:

\begin{itemize}
\item A proposal-free CFNet is proposed to achieve accurate and fast LiDAR panoptic segmentation by solving the bottleneck problems of center modeling and center-based clustering in previous methods.

\item The CFFE is proposed to alleviate the difficulty of modeling the non-existent instance centers and the CDM is designed to efficiently keep one center for each instance.

\item The proposed CFNet is evaluated on the nuScenes and SemanticKITTI LiDAR panoptic segmentation benchmarks. Our CFNet achieves the state-of-the-art performance with a real-time inference speed.

\end{itemize}

\section{Related Work}
\label{sec_related_work}
The LiDAR panoptic segmentation methods usually adopt the LiDAR semantic segmentation networks as the backbone and jointly optimize the semantic segmentation and instance segmentation tasks. The results of the two tasks are fused to generate the final panoptic segmentation results. The LiDAR semantic segmentation backbone is first introduced. Then, the panoptic segmentation methods, achieving instance segmentation upon the semantic segmentation backbone, are grouped into the proposal-based and proposal-free methods, and are further discussed.

\begin{figure*}[!t]
	\centering
	\includegraphics[width=0.95\linewidth]{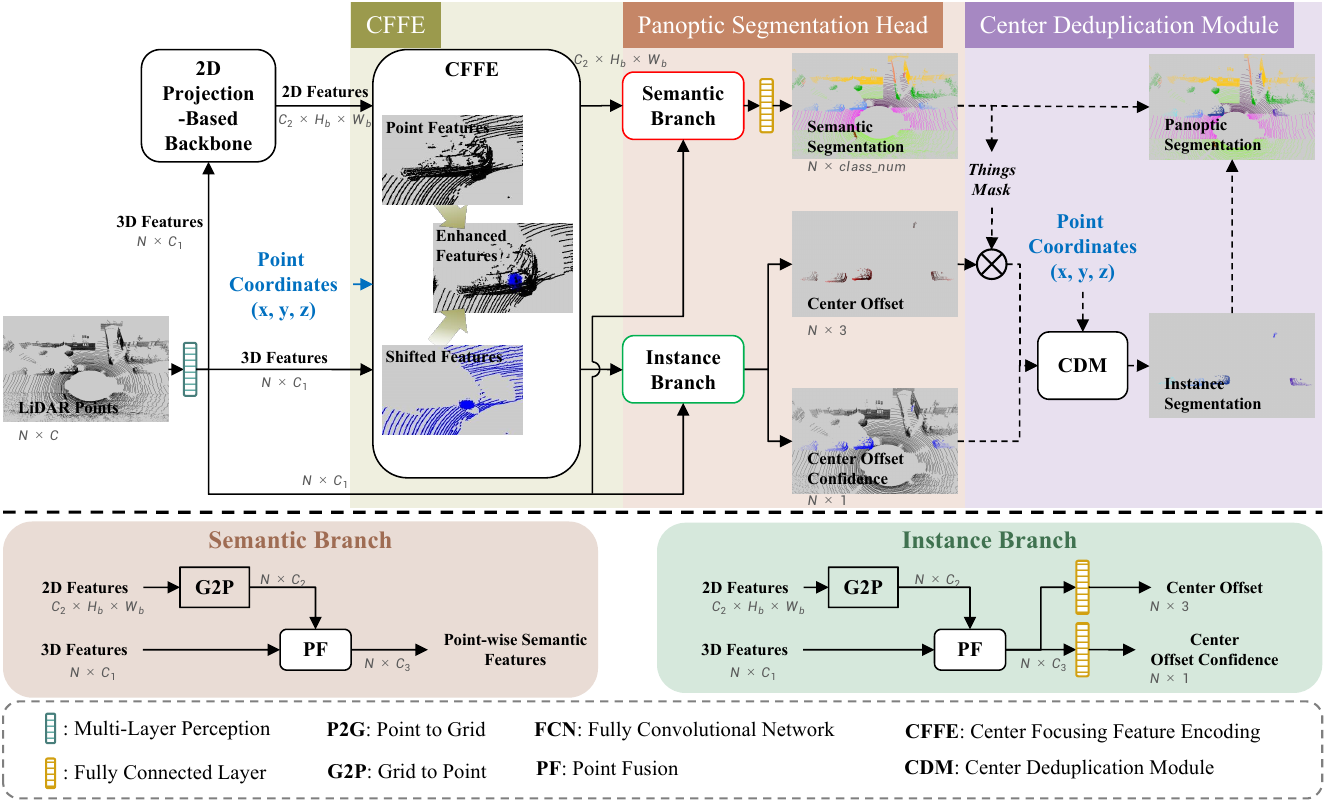}
	\caption{
		The overview of our CFNet. It consists of four steps: 1) the 2D projection-based backbone extracts features on the 2D space; 2) the proposed center focusing feature encoding (CFFE) mimics and enhances the non-existent instance center features; 3) the panoptic segmentation head predicts the output results; 4) the proposed center deduplication module (CDM) achieves instance segmentation that is fused to generate the final panoptic segmentation results. The dashed lines indicate that the operations are only used during inference.
	}
	\label{fig_framework}
\end{figure*}

\textbf{LiDAR semantic segmentation backbone.}\quad The backbone network for LiDAR semantic segmentation can be divided into three groups: point-based, voxel-based, and 2D projection-based methods. The point-based methods~\cite{qi2017pointnet,qi2017pointnetplus,thomas2019kpconv,hu2020randla} directly operate on the raw point clouds but are extremely time-consuming due to the expensive local neighbor searching. The voxel-based methods~\cite{choy20194d,tang2020searching,cheng20212,zhu2021cylindrical} discretize the point clouds into structured voxels, where sparse 3D convolutions are applied. These methods are still difficult to meet the real-time applications, although they achieve the highest accuracy. The 2D projection-based methods are more efficient since most of their computation is done in the 2D space, such as range view (RV)~\cite{cortinhal2020salsanext, milioto2019rangenetplus, xu2020squeezesegv3}, bird’s-eye view (BEV)~\cite{lang2019pointpillars}, polar view~\cite{zhang2020polarnet}, and multi-view fusion~\cite{alnaggar2021multi,li2022cpgnet}. For fast LiDAR panoptic segmentation, the Panoptic-PolarNet~\cite{zhou2021panoptic}, SMAC-Seg~\cite{li2021smac}, and LPSAD~\cite{milioto2020lidar} all adopt the 2D projection-based backbone. A recent real-time semantic segmentation method, namely CPGNet~\cite{li2022cpgnet}, is a 2D projection-based one that explores an end-to-end multi-view fusion framework by fusing the features from the point, BEV, and RV.

\textbf{Proposal-based methods.}\quad The proposal-based methods conduct instance segmentation through a two-stage complicated process. They first detect the foreground instances and then refine the instance segmentation results independently within each detected bounding box. Based on the Mask R-CNN~\cite{he2017mask} for instance segmentation, the MOPT~\cite{hurtado2020mopt} and EfficientLPS~\cite{sirohi2021efficientlps} insert a semantic branch to achieve panoptic segmentation on the range view (RV). Recently, LidarMultiNet~\cite{ye2022lidarmultinet} unifies LiDAR-based 3D object detection, semantic segmentation,
and panoptic segmentation in a single framework to reduce the computation cost by sharing a strong voxel-based backbone.

\textbf{Proposal-free methods.}\quad The proposal-free methods usually apply the class-agnostic clustering to the \emph{things} points to conduct instance segmentation. The LPSAD~\cite{milioto2020lidar} clusters points into instances by regressing the center offsets. The Panoster~\cite{gasperini2021panoster} directly predicts the instance IDs from a learning-based clustering module, where the time-consuming DBSCAN~\cite{ester1996density} is used to refine the instance segmentation results. The DS-Net~\cite{hong2021lidar} proposes a learnable dynamic shifting module that adjusts the kernel functions of the MeanShift~\cite{comaniciu2002mean} to handle instances with various sizes. The GP-S3Net~\cite{razani2021gp} constructs a graph convolutional network (GCN) on the over-segmentation clusters to identify instances. The SMAC-Seg~\cite{li2021smac} proposes a sparse multi-directional attention clustering and center-aware repel loss for instance segmentation. These methods adopt the time-consuming clustering methods. To further accelerate the algorithm, recently, Panoptic-PolarNet~\cite{zhou2021panoptic}, SCAN~\cite{xu2022sparse}, and Panoptic-PHNet~\cite{li2022panoptic} adopt the BEV center heatmap and center offsets for instance segmentation. However, the BEV center heatmap is also costly and confuses on $z$-axis.

\begin{figure*}[!t]
	\centering
	\includegraphics[width=\linewidth]{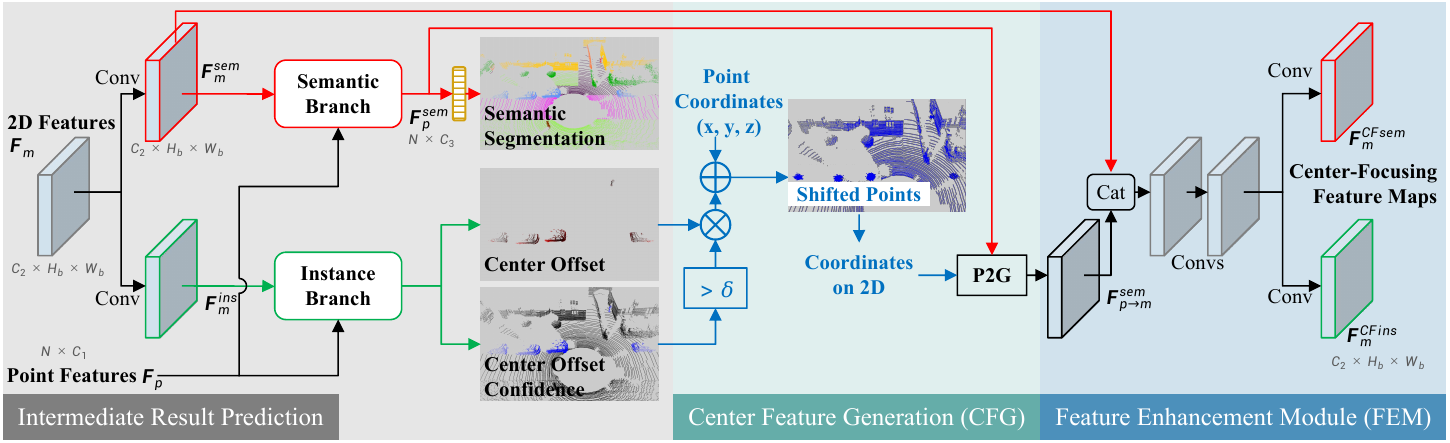}
	\caption{
		The proposed center focusing feature encoding (CFFE). The ``Conv'' represents a 2D convolution with $3 \times 3$ kernels, a batch normalization, and a ReLU layer. The details of the semantic branch and instance branch are shown in Fig.~\ref{fig_framework}. The blue arrows are coordinate-related operations.
	}
	\label{fig_cffe}
\end{figure*}

\section{Approach}
\label{sec_proposed_method}
The input of the LiDAR panoptic segmentation task is the LiDAR point clouds, given by a set of unordered points $\boldsymbol{P} = \left\{(\boldsymbol{p}_i, \boldsymbol{f}_i) \right\}_{i=0}^{N-1}$ (where $\boldsymbol{p}_i=(x,y,z)$ is the 3D coordinates in the cartesian space and $\boldsymbol{f}_i$ denotes additional LiDAR point features, \eg \emph{intensity}). The task aims to assign a set of labels $\boldsymbol{L} = \left\{(s_i, inst_i) \right\}_{i=0}^{N-1}$ to these points, where $s_i$ is the semantic label (\eg \emph{road}, \emph{building}, \emph{car}, \emph{pedestrian}), and $inst_i$ is the instance ID of the $i^{th}$ point. Moreover, $s_i$ can be divided into the uncountable \emph{stuff} classes (\eg \emph{road}, \emph{building}) and countable \emph{things} classes (\eg \emph{car}, \emph{pedestrian}). The instance IDs of \emph{stuff} points are set as 0.

To predict the labels $\boldsymbol{L}$ of the input LiDAR point clouds $\boldsymbol{P}$, our CFNet decomposes this process into four steps, as shown in Fig.~\ref{fig_framework}: 1) an off-the-shelf 2D projection-based backbone is applied to efficiently extract features on the 2D spaces (\eg BEV, RV); 2) a novel center focusing feature encoding (CFFE) is used to generate center-focusing feature maps for more accurate predictions of the instance centers; 3) the panoptic segmentation head fuses the features from the 3D points and 2D spaces to predict the semantic segmentation results, the center offsets, and the confidence scores of the center offsets, respectively; 4) during inference, a post-process is conducted to produce the panoptic segmentation results, where a novel center deduplication module (CDM) operates on the shifted \emph{things} points to select only one center for a single instance.

The design of the backbone in the first step is beyond the scope of the paper, and the last three mentioned steps are illustrated in detail in the following subsections.



\subsection{Center Focusing Feature Encoding}
\label{subsec_cffe}

As mentioned above, the LiDAR points of an object are usually surface-aggregated, especially for \emph{car} and \emph{truck} categories, resulting in the fact that the center of an object is imaginary and does not exist in the LiDAR point clouds. To encode the features of the non-existent centers, a novel center focusing feature encoding (CFFE) is proposed, which takes the 2D features from the backbone and the 3D point coordinates as inputs and generates the enhanced center-focusing feature maps as shown in Fig.~\ref{fig_framework}.

The CFFE module consists of three steps, including intermediate result prediction, center feature generation, and feature enhancement module, as shown in Fig.~\ref{fig_cffe}.

\textbf{Intermediate result prediction.}\quad In this step, the CFFE predicts intermediate results (including the semantic segmentation, center offset, and its confidence scores) according to the 2D features $\boldsymbol{F}_m$ and 3D point features $\boldsymbol{F}_p$ for subsequent center feature simulation. Specifically, two convolution layers are applied on the 2D features $\boldsymbol{F}_m$ independently, to generate semantic features $\boldsymbol{F}_m^{sem}$ and instance features $\boldsymbol{F}_m^{ins}$ ($m$ is the specific 2D view, such as RV~\cite{cortinhal2020salsanext, milioto2019rangenetplus, xu2020squeezesegv3}, BEV~\cite{lang2019pointpillars}, and polar view~\cite{zhang2020polarnet}.)
\begin{align}
	\boldsymbol{F}_m^{sem} = Conv(\boldsymbol{F}_m; \boldsymbol{\theta}_1),\\
	\boldsymbol{F}_m^{ins} = Conv(\boldsymbol{F}_m; \boldsymbol{\theta}_2),
\end{align}where $Conv$ denotes sequential 2D convolution, batch normalization, and ReLU operations, and $\boldsymbol{\theta}_1$ and $\boldsymbol{\theta}_2$ are their learnable parameters. Then, the semantic branch generates per-point 3D semantic features $\boldsymbol{F}_p^{sem}$ by fusing the point features $\boldsymbol{F}_p$ and 2D semantic features $\boldsymbol{F}_m^{sem}$,
\begin{equation}
	\boldsymbol{F}_p^{sem} = Seg(\boldsymbol{F}_p, \boldsymbol{F}_m^{sem}; \boldsymbol{\theta}_3),\\
\end{equation}where $Seg$ is the semantic branch, and $\boldsymbol{\theta}_3$ is the parameters.

Finally, the intermediate semantic results $\hat{\boldsymbol{S'}}=\{\hat{s'}_i\}_{i=0}^{N-1}$ is produced based on $\boldsymbol{F}_p^{sem}$. The intermediate results of center offsets $\hat{\boldsymbol{O'}}=\{\hat{\boldsymbol{o'}}_i\}_{i=0}^{N-1}$ and confidence scores $\boldsymbol{\hat{C'}}=\{\hat{c'}_i\}_{i=0}^{N-1}$ are predicted by the instance branch with the point features $\boldsymbol{F}_p$ and 2D instance features $\boldsymbol{F}_m^{ins}$,
\begin{align}
	\hat{\boldsymbol{S'}} &= FC(\boldsymbol{F}_p^{sem}; \boldsymbol{\theta}_4),\\
	\hat{\boldsymbol{O'}},  \boldsymbol{\hat{C'}} &= Ins(\boldsymbol{F}_p, \boldsymbol{F}_m^{ins}; \boldsymbol{\theta}_5),
\end{align}where $Ins$ is the instance branch, and $FC$ represents the fully-connected layer.
The structure and the training objective of semantic and instance branches are the same as those in the panoptic segmentation head and are elucidated in Fig.~\ref{fig_framework} and section~\ref{subsec_head}.

\textbf{Center feature generation (CFG).}\quad In this step, the CFFE generates the shifted center features $\boldsymbol{F}_{p \rightarrow m}^{sem}$  by shifting the 3D semantic point features $\boldsymbol{F}_p^{sem}$ to the predicted centers according to the mentioned intermediate results. First, the coordinates of a predicted center $\boldsymbol{p}^{shift}_i$ are computed by,
\begin{equation}
\boldsymbol{p}^{shift}_i = \boldsymbol{p} _i+ \hat{\boldsymbol{o'}}_i * \mathbbm{1}[\hat{c'}_i > \delta],
\end{equation}
where $\boldsymbol{p}_i$ is the original 3D LiDAR point coordinates, and $\mathbbm{1}[\hat{c'}_i > \delta]$ is a binary indicator of whether the confidence $\hat{c'}_i$ is greater than $\delta=0.2$. In other words, it does not shift the \emph{stuff} points or the \emph{things} points with low confidence scores.

Then, the shifted 3D points $\{\boldsymbol{p}^{shift}_i\}_{i=0}^{N-1}$ are served as the new coordinates of the feature points, and the Point to Grid (P2G) operation in \cite{li2022cpgnet} re-projects the 3D semantic features $\boldsymbol{F}_p^{sem}$ onto the 2D projected feature maps $\boldsymbol{F}_{p \rightarrow m}^{sem}$ with this new coordinates,
\begin{equation}
	\boldsymbol{F}_{p \rightarrow m}^{sem} = P2G(\boldsymbol{F}_p^{sem}; \{\boldsymbol{p}^{shift}_i\}_{i=0}^{N-1}).
\end{equation}Compared with $\boldsymbol{F}_p^{sem}$, the feature maps $\boldsymbol{F}_{p \rightarrow m}^{sem}$ pay more attention to the imaginary centers since most of the \emph{things} points have been shifted to their predicted centers.

\textbf{Feature enhancement module (FEM).}\quad The CFFE finally fuses the semantic feature maps $\boldsymbol{F}_m^{sem}$ and the re-projected shifted center feature maps $\boldsymbol{F}_{p \rightarrow m}^{sem}$ to generate center-focusing semantic feature maps $\boldsymbol{F}_m^{CFsem}$ and instance feature maps $\boldsymbol{F}_m^{CFins}$, which are used by the subsequent semantic and instance branches for more accurate predictions. The enhancement module is composed of a simple concatenation operation and several convolution layers, whose details are in the supplementary material.

In addition, for application of the multi-view fusion backbone (\eg ~\cite{li2022cpgnet} shown in the supplementary material), the feature maps $\boldsymbol{F}_m$, $\boldsymbol{F}_m^{sem}$, $\boldsymbol{F}_m^{ins}$, $\boldsymbol{F}_{p \rightarrow m}^{sem}$ (\eg $m\in \{\text{RV}, \text{BEV}\}$) of each view are computed independently according to the above process. Then, they are fused in the point fusion (PF) module~\cite{li2022cpgnet} to generate integrated 3D point features and per-point prediction.

\subsection{Panoptic Segmentation Head}
\label{subsec_head}
For better modeling the instance center, the panoptic segmentation head uses a semantic branch to predict the semantic segmentation, and an instance branch to estimate both the center offsets and the newly introduced confidence scores, given the center-focusing semantic and instance feature maps. The architectures of both branches follow the segmentation head in \cite{li2022cpgnet} and are briefly introduced, while their training objectives are elaborated on.

\textbf{Semantic branch.}\quad For a per-point prediction, the semantic branch first applies the Grid to Point (G2P) operations to acquire the 3D point-wise features from the 2D semantic feature maps. Then, a PF module fuses the point-wise features from the G2P operation and the original 3D points to generate point-wise semantic features. The details of the G2P and PF can be found in \cite{li2022cpgnet}. After obtaining the point-wise semantic features, a fully connected (FC) layer is used to predict the final per-point semantic result $\hat{\psi}_i^j$ ($\hat{\psi}_i^j\in [0,1]$). $\hat{\psi}_i^j$ represents the probability of the $i^{th}$ LiDAR point belonging to the $j^{th}$ class. The predicted semantic label $\hat{s}_i$ is acquired by selecting the most probable class $\hat{s}_i = \mathop{\arg\max}\nolimits_{j} \hat{\psi}_i^j$.

Referring to the CPGNet~\cite{li2022cpgnet}, the same loss functions are adopted, including weighted cross entropy (WCE) loss $\mathcal{L}_{wce}$, Lov{\'a}sz-Softmax loss $\mathcal{L}_{ls}$~\cite{berman2018lovasz}, and transformation consistency loss $\mathcal{L}_{tc}$.

\textbf{Instance branch.}\quad Similar to the semantic branch, the instance branch also adopts the G2P operation and a PF module to acquire point-wise instance features. An FC layer is applied to predict the per-point center offsets $\hat{\boldsymbol{o}}_i$. The ground-truth for $\hat{\boldsymbol{o}}_i$ is the offset vector $\boldsymbol{e}_i - \boldsymbol{p}_i$ from the $i^{th}$ point $\boldsymbol{p}_i$ to its corresponding instance center $\boldsymbol{e}_i$.

For center offset regression, the loss function for optimizing $\hat{\boldsymbol{o}}_i$ only considers the \emph{things} classes and is formulated as follows,
\begin{equation}
\label{eq_offset}
\mathcal{L}_{\boldsymbol{o}_i}=
\begin{cases}
||\hat{\boldsymbol{o}}_i - (\boldsymbol{e}_i - \boldsymbol{p}_i)||_2 & \text{if}~s_i \in \emph{things}~\text{classes},\\
0 & \text{otherwise} ,
\end{cases}
\end{equation}
where $\boldsymbol{e}_i$ is the axis-aligned center of the instance~\cite{hong2021lidar}. Then, the losses are summarized as the following,
\begin{equation}
	\mathcal{L}_{o} = \frac{1}{N^{things}} \sum_{i=0}^{N-1} \mathcal{L}_{\boldsymbol{o}_i},
\end{equation}
where $N$ and $N^{things}$ are the numbers of all points and \emph{things} points, respectively.

For confidence score regression, another FC layer is used to predict the per-point confidence score $\hat{c}_i$ to indicate the accuracy degree of $\hat{\boldsymbol{o}}_i$. $\hat{c}_i$ is activated by a sigmoid function to ensure $\hat{c}_i \in [0,1]$. The ground-truth label $c_i$ for supervising $\hat{c}_i$ is generated by
\begin{equation}
	\label{eq_conf_gt}
	c_i=
	\begin{cases}
		\exp(-\frac{\mathcal{L}_{\boldsymbol{o}_i}^2}{2\sigma^2}) & \text{if}~s_i \in \emph{things}~\text{classes},\\
		0 & \text{otherwise}.
	\end{cases}
\end{equation}
For the \emph{things} points, the lower $\mathcal{L}_{\boldsymbol{o}_i}$ is, the higher $c_i$ is. It means that the point with a more accurate regression of the center offset has a higher confidence score.

The weighted binary cross entropy loss $\mathcal{L}_{wbce}$ is applied,
\begin{equation}
	\mathcal{L}_{wbce} = -\frac{1}{N} \sum_{i=0}^{N-1} 6c_i\log(\hat{c}_i) + (1 - c_i)\log(1 - \hat{c}_i),
\end{equation}
where the \emph{things} points are manually emphasized since the number of them is much less than that of the \emph{stuff} points.

Finally, the loss for each group of results (from the CFNet or CFFE) is defined as
\begin{align}
	\mathcal{L} &= \mathcal{L}_{wce} + 3\mathcal{L}_{ls} + \mathcal{L}_{tc} + 2\mathcal{L}_{o} + \mathcal{L}_{wbce}.
\end{align}
The total loss is the sum of the two losses sourced from the CFNet and CFFE.

\subsection{Center Deduplication Module}
\label{subsec_cdm}

Given the final predictions of the semantic segmentation $\hat{\boldsymbol{S}}$, center offsets $\hat{\boldsymbol{O}}$, and confidence scores $\boldsymbol{\hat{C}}$, this section introduces how to exploit the detected centers to acquire the panoptic segmentation results during inference, and illustrates the key module, center deduplication module (CDM).

\textbf{Post-process.}\quad For the panoptic segmentation results, instance segmentation is first generated and then the final panoptic segmentation is acquired by fusing the semantic and instance segmentation labels. There are five steps to achieve the final panoptic segmentation:

 1) The \emph{things} points $\{\boldsymbol{\tilde{p}}_t \}_{t=0}^{M-1} = \{\boldsymbol{p}_i | \hat{s}_i\in \text{\emph{things} classes}\}$ are selected according to the predicted semantic labels, with their offsets $\{\boldsymbol{\tilde{o}}_t \}_{t=0}^{M-1}$ and confidence scores $\{\tilde{c}_t \}_{t=0}^{M-1}$ (where $M$ is the number of \emph{things} points).
 
 2) Each shifted \emph{things} point $\boldsymbol{\tilde{p}}^{shift}_t$ is given by $\boldsymbol{\tilde{p}}^{shift}_t = \boldsymbol{\tilde{p}}_t + {\boldsymbol{\tilde{o}}}_t$ and is severing as an instance center candidate.
 
 3) A center $\tilde{\boldsymbol{e}}_j$ for each instance $j$ is selected by the CDM based on the coordinates $\boldsymbol{\tilde{p}}^{shift}_t$ and its confidence score $\tilde{c}_t$, while other candidates are suppressed.
 
 4) The instance ID $\widehat{inst}_t$ is acquired by assigning the shifted \emph{things} points $\boldsymbol{\tilde{p}}^{shift}_t$ to the closest center in all centers $\{\tilde{\boldsymbol{e}}_j\}_{j=0}^{D-1}$ ($D$ is the number of detected instances).
 
 5) The majority voting~\cite{zhou2021panoptic, hong2021lidar} reassigns the most frequent semantic label to all points of a predicted instance to further guarantee the consistency of semantic labels within a predicted instance.

\textbf{Center deduplication module (CDM).}\quad The CDM takes the shifted points $\{\boldsymbol{\tilde{p}}^{shift}_t\}_{t=0}^{M-1}$ and confidence scores $\{\tilde{c}_t\}_{t=0}^{M-1}$ as input to get an instance center $\hat{\boldsymbol{e}}_j$ for each instance $j$. Inspired by the bounding box NMS, our CDM suppresses the candidate centers with lower scores within a euclidean distance threshold $d$. The pseudo-code of our CDM is shown in Algorithm~\ref{algo_cdm}, where two centers with a distance less than $d$ are considered as the same instance. The process of our CDM is simple and can be easily implemented in CUDA.

\begin{algorithm}[t]
 \caption{Center deduplication module}
 \label{algo_cdm}
    
 \KwIn{Shifted \emph{things} points $\{\boldsymbol{\tilde{p}}^{shift}_t \in \mathbb{R}^{3}\}_{t=0}^{M-1}$,  confidence scores $\{\tilde{c}_t  \in [0,1]\}_{t=0}^{M-1}$,\\
  \qquad\quad distance threshold $d \in \mathbb{R}$}
 \KwOut{Instance centers $\{\tilde{\boldsymbol{e}}_j\in \mathbb{R}^{3}\}_{j=0}^{D-1}$}
 \BlankLine
 $order$ = sort\_index($\{\tilde{c}_t\}_{t=0}^{M-1}$, decending=true)\\
 initialize a boolean vector $rej \in \mathbb{B}^{M}$ with all elements of false and an empty list $keep$\\
 \For{$k \leftarrow 0$ \KwTo $M-1$}{
  \If{{\bf not} $rej[order[k]]$}{
   $keep.append(order[k])$\\
   \For{$h \leftarrow k+1$ \KwTo $M-1$}{
    $dist = ||\boldsymbol{\tilde{p}}^{shift}_{order[k]} - \boldsymbol{\tilde{p}}^{shift}_{order[h]}||_2$\\
    \If{$dist < d$}{
     $rej[order[h]] = \text{true}$
    }
   }
  }
 }
 $\{\tilde{\boldsymbol{e}}_j\} = \boldsymbol{\tilde{p}}^{shift}_{keep} $

\end{algorithm}

\section{Experiments}
\label{sec_experiments}
The proposed CFNet is compared with the existing methods on the public SemanticKITTI~\cite{behley2019semantickitti} and nuScenes~\cite{caesar2020nuscenes} panoptic segmentation benchmarks.

\subsection{Experimental Setup}
\label{subsec_exp_setup}

\textbf{Datasets.}\quad The SemanticKITTI~\cite{behley2019semantickitti} consists of 22 sequences collected by a Velodyne HDL-64E $360^\circ$ rotating LiDAR with 64 beams vertically. It contains 43,552 LiDAR scans in total and is split into a training set with 19,130 scans from sequences 00 to 10 except 08, a validation set containing sequence 08 with 4,071 scans, and a test set including the rest sequences with 20,351 scans. The test set is only provided with point clouds and used for the online leaderboards. For the panoptic segmentation, it has 19 valid categories, including 11 \emph{stuff} and 8 \emph{things} categories.

The nuScenes~\cite{caesar2020nuscenes} is a newly released benchmark with 1,000 scenes collected by a Velodyne HDL-32E $360^\circ$ rotating LiDAR with 32 beams vertically. The dataset is collected in Boston and Singapore. It uses 28,130 scans for training, 6,019 for validation, and 6,008 for testing. The panoptic segmentation benchmark provides 16 categories, including 6 \emph{stuff} and 10 \emph{things} categories~\cite{fong2022panoptic}.

\textbf{Evaluation metric.}\quad As the official benchmarks~\cite{behley2021benchmark, fong2022panoptic} suggest, the panoptic quality (PQ) is adopted to evaluate the performance of LiDAR panoptic segmentation. The PQ~\cite{behley2021benchmark, fong2022panoptic} measures the overall quality of the panoptic segmentation, and can be decomposed into two terms: 1) the segmentation quality (SQ) is the average IoU of all matched pairs; 2) the recognition quality (RQ) measures the $F_1$ score. These three metrics are also reported separately on the \emph{stuff} and \emph{things} classes, including PQ$^{St}$, SQ$^{St}$, RQ$^{St}$, and PQ$^{Th}$, SQ$^{Th}$, RQ$^{Th}$, respectively. The PQ$^{\dag}$ ~\cite{porzi2019seamless} is also reported by replacing the PQ of each \emph{stuff} class with its IoU. For evaluating the sub-task of semantic segmentation, mean Intersection over Union (mIoU)~\cite{behley2019semantickitti} is reported.

\textbf{Training details.}\quad In the experiments, two 2D projection-based backbones, namely PolarNet~\cite{zhang2020polarnet} and CPGNet~\cite{li2022cpgnet}, are used in our CFNet. For the backbone of PolarNet~\cite{zhang2020polarnet}, the training schedules and data augmentation are the same as those of Panoptic-PolarNet~\cite{zhou2021panoptic} for a fair comparison. Besides, for efficiency, only one stage version of CPGNet~\cite{li2022cpgnet} is used in our CFNet. It is trained from scratch for 48 epochs with a batch size of 8 and takes around 24 hours on 8 NVIDIA RTX 3090 GPUs. Stochastic gradient descent (SGD) is used as the optimizer, where the initial learning rate, weight decay, and momentum are 0.02, 0.001, and 0.9, respectively. The learning rate is decayed by 0.1 per 10 epochs. In training, the data augmentation is applied as a convention, including random flipping along the $x$ and $y$ axes, random global scale sampled from $[0.95, 1.05]$, random rotation around the $z$-axis, and random Gaussian noise $\mathcal{N}(0, 0.02)$.

\begin{table*}[!t]
\centering
\resizebox{0.9\linewidth}{!}{
\begin{tabular}{c|l|cc|c|ccccc|c}

\hline

& \multirow{2}{*}{Backbone} & \multicolumn{2}{c|}{CFFE} & \multirow{2}{*}{Deduplication} & \multirow{2}{*}{PQ} & \multirow{2}{*}{PQ$^\dag$} & \multirow{2}{*}{PQ$^{Th}$} & \multirow{2}{*}{PQ$^{St}$} & \multirow{2}{*}{mIoU} & \multirow{2}{*}{RT(ms)}\\

\cline{3-4}

 & & CFG & FEM & & & & & & & \\

\hline \hline

a & \multirow{4}{*}{PolarNet~\cite{zhang2020polarnet}} & & & BEV Center Heatmap~\cite{zhou2021panoptic} & 59.1 & 64.1 & 65.7 & 54.3 & 64.5 & 65.3 + 20.9\\

b & & & \checkmark & BEV Center Heatmap~\cite{zhou2021panoptic} & 59.5 & 64.3 & 66.2 & 54.4 & 64.6 & 69.7 + 20.9\\

c & & \checkmark & \checkmark & BEV Center Heatmap~\cite{zhou2021panoptic} & 60.4 & 65.6 & 67.2 & 55.1 & 65.2 & 71.9 + 20.9\\

d & & \checkmark & \checkmark & CDM~[Ours] & {\bf 60.6} & {\bf 65.7} & {\bf 67.8} & {\bf 55.2} & {\bf 65.4} & {\bf 71.9 + 3.6}\\

\hline

e & \multirow{6}{*}{CPGNet~\cite{li2022cpgnet}} & & & DBSCAN~\cite{ester1996density} & 59.5 & 64.1 & 63.8 & 56.3 & 64.6 & 31.6+24.9\\

f & & & & HDBSCAN~\cite{campello2013density} & 58.3 & 62.9 & 60.9 & 56.3 & 64.9 & 31.6+48.7\\

g & & & & MeanShift~\cite{comaniciu2002mean} & 59.9 & 64.5 & 64.7 & 56.3 & 64.9 & 31.6+84.6\\

h & & & & CDM~[Ours] & 60.5 & 65.6 & 66.1 & 56.5 & 66.3 & {\bf 32.7+2.3}\\

i & & & \checkmark & CDM~[Ours] & 60.8 & 65.9 & 66.3 & 56.9 & 66.5 & 37.4+2.3\\

j & & \checkmark & \checkmark & CDM~[Ours] & {\bf 62.7} & {\bf 67.5} & {\bf 70.0} & {\bf 57.3} & {\bf 67.4} & 41.2+2.3\\

\hline
\end{tabular}
}
\caption{Ablation studies on the SemanticKITTI validation set. RT: running time.}
\label{table_ab_semkitti}
\end{table*}

\subsection{Ablation Studies}

Ablative analyses are conducted on the SemanticKITTI validation set with the above experimental setup. As shown in Table~\ref{table_ab_semkitti}, two 2D projection-based backbones, namely PolarNet~\cite{zhang2020polarnet} and CPGNet~\cite{li2022cpgnet}, are incorporated with our framework to validate the proposed CFFE and CDM. Row (a) shows the original Panoptic-Polarnet~\cite{zhou2021panoptic}, while rows (b,c,d) are implemented based on the official code of Panoptic-Polarnet~\cite{zhou2021panoptic}. The running time is reported by the time of the model inference plus that of the post-process.

\textbf{Effects of the CFFE.}\quad The CFFE is configured in three ways: 1) the CFFE is not applied (a,e,f,g,h); 2) the center feature generation (CFG) in the CFFE is removed (b,i), which is equal to the setting of only imposing the intermediate supervision; 3) the whole CFFE is applied (c,d,j). For the two backbones, a common observation is that the improvements primarily come from the CFG rather than the intermediate supervision, especially for the \emph{things} classes (a,b,c and h,i,j). In terms of the running time, The CFFE brings acceptable latency of +6.6\,ms and +8.5\,ms for these two backbones, respectively.

Moreover, the errors in predicting \emph{things} center offsets are calculated for both the CFNet with CFFE and its intermediate results as shown in Table~\ref{table_error}. Equipped with the backbone of CPGNet~\cite{li2022cpgnet}, the proposed CFFE reduces the errors of predicting the \emph{things} center offsets. The observations demonstrate that explicitly modeling center features helps in identifying an instance.


\begin{table}[!b]
\centering
\setlength\tabcolsep{3pt}
\resizebox{0.7\linewidth}{!}{
	\begin{tabular}{c|c|c}
	\hline
	&  training set & validation set \\
	\hline\hline
	Intermediate & 1.35 & 1.22 \\
	\hline
	CFNet  & {\bf 1.19} & {\bf 1.13} \\
	\hline
	\end{tabular}
}
\caption{Mean errors measured in meter (m) of \emph{things} center offsets for the intermediate results and the CFNet with the CFFE on the SemanticKITTI training and validation set.}
\label{table_error}
\end{table}

\textbf{Effects of the CDM.}\quad For the proposal-free methods, there are primarily three ways to identify each instance: 1) the class-agnostic clustering methods (\eg DBSCAN, HDBSCAN, MeanShift) (e,f,g), 2) the BEV center heatmap~\cite{zhou2021panoptic} (a,b,c), and 3) our CDM (d,h,i,j). For the backbone of PolarNet~\cite{zhang2020polarnet}, our CDM outperforms the BEV center heatmap~\cite{zhou2021panoptic} by +0.2 for the PQ (c,d), while our CDM runs faster. For the backbone of CPGNet~\cite{li2022cpgnet}, compared with the class-agnostic clustering methods, our CDM also achieves the best performance on both accuracy (+0.6$\sim$+2.2 for the PQ) and running time (-21.5\,ms$\sim$-81.2\,ms), though it slows down the model inference by +1.1\,ms (g,h) owing to the additional confidence prediction.

More ablation studies can be seen in the supplementary material.



\subsection{Evaluation Comparisons}

\textbf{Quantitative results.}\quad In Table~\ref{table_sk} and Table~\ref{table_nusc}, our CFNet with CPGNet as its backbone is compared with the existing methods on the SemanticKITTI test set and nuScenes validation set, respectively. From top to bottom, the methods are grouped as proposal-based and proposal-free methods. It is found that our proposal-free CFNet surpasses the previous methods by a large margin on most metrics. Specifically, on the SemanticKITTI test set, our CFNet outperforms the previous state-of-the-art SCAN~\cite{xu2022sparse} and Panoptic-PHNet~\cite{li2022panoptic} by +1.9 for the PQ, and +1.5 for the RQ. On the nuScenes validation set, our CFNet outperforms the most competitive Panoptic-PHNet~\cite{li2022panoptic} by +0.4 for the PQ, +0.6 for the SQ, and +0.4 for the RQ, respectively.

\begin{table*}[!ht]
\centering
\setlength\tabcolsep{3pt}
\resizebox{0.9\linewidth}{!}{
\begin{tabular}{l|cccc|ccc|ccc|c|c}

\hline

Methods & PQ & PQ$^\dag$ & SQ & RQ & PQ$^{Th}$ & SQ$^{Th}$ & RQ$^{Th}$ & PQ$^{St}$ & SQ$^{St}$ & RQ$^{St}$ & mIoU & RT(ms)\\
\hline \hline

{\bf Proposal-based Methods} & & & & & & & & & & & & \\
RangeNet++~\cite{milioto2019rangenetplus}\&PointPillars~\cite{lang2019pointpillars} & 37.1 & 45.9 & 75.9 & 47.0 & 20.2 & 75.2 & 25.2 & 49.3 & 76.5 & 62.8 & 52.4 & 416.6\\

KPConv~\cite{thomas2019kpconv}\&PointPillars~\cite{lang2019pointpillars} & 44.5 & 52.5 & 80.0 & 54.4 & 32.7 & 81.5 & 38.7 & 53.1 & 79.0 & 65.9 & 58.8 & 526.3\\

PanopticTrackNet~\cite{hurtado2020mopt} & 43.1 & 50.7 & 78.8 & 53.9 & 28.6 & 80.4 & 35.5 & 53.6 & 77.7 & 67.3 & 52.6 & 147\\

EfficientLPS~\cite{sirohi2021efficientlps} & 57.4 & 63.2 & 83.0 & 68.7 & 53.1 & 87.8 & 60.5 & 60.5 & 79.5 & {\bf 74.6} & 61.4 & -\\

\hline \hline
{\bf Proposal-free Methods} & & & & & & & & & & & & \\
LPSAD~\cite{milioto2020lidar} & 38.0 & 47.0 & 76.5 & 48.2 & 25.6 & 76.8 & 31.8 & 47.1 & 76.2 & 60.1 & 50.9 & 84.7\\

DS-Net~\cite{hong2021lidar} & 55.9 & 62.5 & 82.3 & 66.7 & 55.1 & 87.2 & 62.8 & 56.5 & 78.7 & 69.5 & 61.6 & 294.1\\

Panoster~\cite{gasperini2021panoster} & 52.7 & 59.9 & 80.7 & 64.1 & 49.4 & 83.3 & 58.5 & 55.1 & 78.8 & 68.2 & 59.9 & -\\

GP-S3Net~\cite{razani2021gp} & 60.0 & 69.0 & 82.0 & 72.1 & 65.0 & 86.6 & {\bf 74.5} & 56.4 & 78.7 & 70.4 & {\bf 70.8} & 270.3\\

SMAC-Seg~\cite{li2021smac} & 56.1 & 62.5 & 82.0 & 67.9 & 53.0 & 85.6 & 61.8 & 58.4 & 79.3 & 72.3 & 63.3 & 99\\

Panoptic-PolarNet~\cite{zhou2021panoptic} & 54.1 & 60.7 & 81.4 & 65.0 & 53.3 & 87.2 & 60.6 & 54.8 & 77.2 & 68.1 & 59.5 & 86.2\\

SCAN~\cite{xu2022sparse} & 61.5 & 67.5 & 84.5 & 72.1 & 61.4 & 88.1 & 69.3 & {\bf 61.5} & 81.8 & 74.1 & 67.7 & 78.1\\

Panoptic-PHNet~\cite{li2022panoptic} & 61.5 & 67.9 & 84.8 & 72.1 & 63.8 & {\bf 90.7} & 70.4 & 59.9 & 80.5 & 73.3 & 66.0 & 69.3\\

\hline
{\bf CFNet~[Ours] w/CPGNet~\cite{li2022cpgnet}} & {\bf 63.4} & {\bf 69.7} & {\bf 85.2} & {\bf 73.6} & {\bf 66.7} & 89.8 & 74.3 & 61.0 & {\bf 81.9} & 73.1 & 68.3 & {\bf 41.2 + 2.3}\\

\hline
\end{tabular}
}
\caption{Comparison results on the SemanticKITTI test set.}
\label{table_sk}
\end{table*}

\begin{table*}[!ht]
\centering
\resizebox{0.9\linewidth}{!}{
\begin{tabular}{l|cccc|ccc|ccc|c}

\hline

Methods & PQ & PQ$^\dag$ & SQ & RQ & PQ$^{Th}$ & SQ$^{Th}$ & RQ$^{Th}$ & PQ$^{St}$ & SQ$^{St}$ & RQ$^{St}$ & mIoU \\
\hline \hline

{\bf Proposal-based Methods} & & & & & & & & & & & \\

Cylinder3D~\cite{zhu2021cylindrical}\&PointPillars~\cite{lang2019pointpillars} & 36.0 & 44.5 & 83.3 & 43.0 & 23.3 & 83.7 & 27.0 & 57.2 & 82.7 & 69.6 & 52.3\\

Cylinder3D~\cite{zhu2021cylindrical}\&SECOND~\cite{yan2018second} & 40.1 & 48.4 & 84.2 & 47.3 & 29.0 & 84.4 & 33.6 & 58.5 & 83.7 & 70.1 & 58.5\\

PanopticTrackNet~\cite{hurtado2020mopt} & 51.4 & 56.2 & 80.2 & 63.3 & 45.8 & 81.4 & 55.9 & 60.4 & 78.3 & 75.5 & 58.0\\

EfficientLPS~\cite{sirohi2021efficientlps} & 62.0 & 65.6 & 83.4 & 73.9 & 56.8 & 83.2 & 68.0 & 70.6 & 83.8 & 83.6 & 65.6\\

\hline \hline
{\bf Proposal-free Methods} & & & & & & & & & & & \\
LPSAD~\cite{milioto2020lidar} & 50.4 & 57.7 & 79.4 & 62.4 & 43.2 & 80.2 & 53.2 & 57.5 & 78.5 & 71.7 & 62.5\\

DS-Net~\cite{hong2021lidar} & 42.5 & 51.0 & 83.6 & 50.3 & 32.5 & 83.1 & 38.3 & 59.2 & 84.4 & 70.3 & 70.7\\

GP-S3Net~\cite{razani2021gp} & 61.0 & 67.5 & 84.1 & 72.0 & 56.0 & 85.3 & 65.2 & 66.0 & 82.9 & 78.7 & 75.8\\

SMAC-Seg~\cite{li2021smac} & 67.0 & 71.8 & 85.0 & 78.2 & 65.2 & 87.1 & 74.2 & 68.8 & 82.9 & 82.2 & 72.2\\

Panoptic-PolarNet~\cite{zhou2021panoptic} & 63.4 & 67.2 & 83.9 & 75.3 & 59.2 & 84.1 & 70.3 & 70.4 & 83.6 & 83.5 & 66.9\\

SCAN~\cite{xu2022sparse} & 65.1 & 68.9 & 85.7 & 75.3 & 60.6 & 85.7 & 70.2 & 72.5 & 85.7 & 83.8 & 77.4\\

Panoptic-PHNet~\cite{li2022panoptic} & 74.7 & 77.7 & 88.2 & 84.2 & 74.0 & 89.0 & 82.5 & 75.9 & 86.8 & 86.9 & {\bf 79.7}\\

\hline
{\bf CFNet~[Ours] w/CPGNet~\cite{li2022cpgnet}} & {\bf 75.1} & {\bf 78.0} & {\bf 88.8} & {\bf 84.6} & {\bf 74.8} & {\bf 89.8} & {\bf 82.9} & {\bf 76.6} & {\bf 87.1} & {\bf 87.3} & 79.3\\
\hline
\end{tabular}
}
\caption{Comparison results on the nuScenes validation set.}
\label{table_nusc}
\end{table*}

\textbf{Running time.}\quad The running time is reported on the SemanticKITTI. It is measured with PyTorch FP32 without any model acceleration tricks on a single NVIDIA RTX 3090 GPU. For clarity, the running time of our method is further divided into two parts, including the time costs of the model and post-process. Based on the same kind of devices as the existing methods, our CFNet runs 43.5\,ms (41.2\,ms for model inference and 2.3\,ms for post-process) and is 1.6 times faster than the most efficient Panoptic-PHNet~\cite{li2022panoptic}.

\begin{figure}[!ht]
	\centering
	\includegraphics[width=0.95\linewidth]{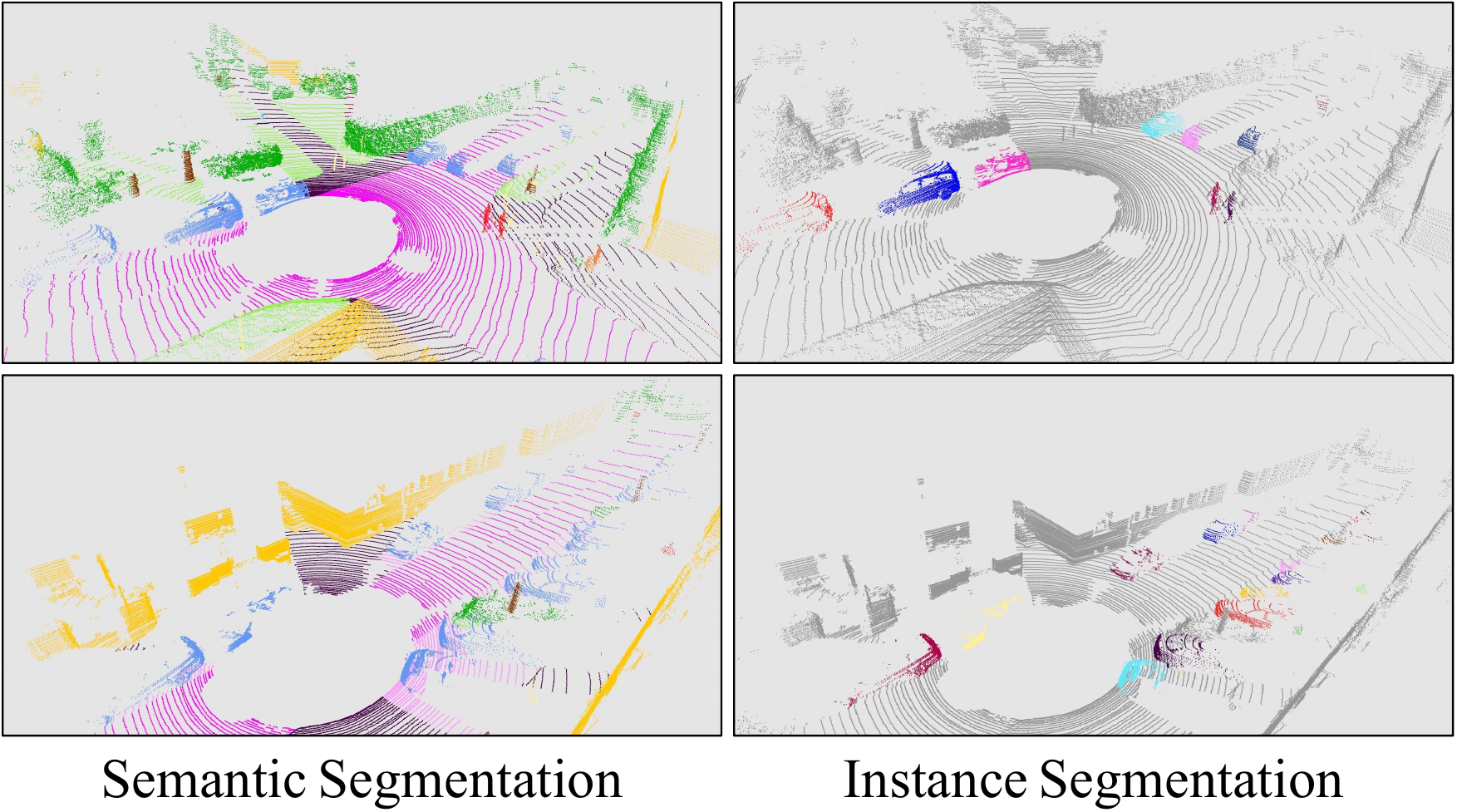}
	\caption{
		Visualizations of our CFNet on the SemanticKITTI test set. Different colors represent different classes or instances.
	}
	\label{fig_vis}
\end{figure}

\textbf{Qualitative results.}\quad The visualization results are shown in Fig.~\ref{fig_vis}. Our CFNet can distinguish adjacent pedestrians or cars. Besides, the boundaries of instances can be accurately segmented. More qualitative results can be seen in the supplementary material.

\section{Conclusion}
A novel proposal-free CFNet is proposed for real-time LiDAR panoptic segmentation. For better modeling and exploiting the non-existent instance centers, a novel CFFE is proposed to generate enhanced center-focusing feature maps and a CDM is introduced to keep only one center for each instance and then assign the shifted \emph{things} points to the closest center for acquiring instance IDs. From the experiments, it can be seen that center modeling and exploiting is a key problem in the proposal-free LiDAR panoptic segmentation methods, and mimicking the non-existent center features is promising and shows a clear benefit.

\section*{Acknowledgements}
The research project is partially supported by the National Key R\&D Program of China (No.2021ZD0111902), and the National Natural Science Foundation of China (No.U21B2038, 62272015, U19B2039).

{\small
\bibliographystyle{ieee_fullname}
\bibliography{egbib}
}

\newpage
\section*{\LARGE{Supplementary Material}}
\appendix

In the supplementary material, the implementation details, extensive ablation studies, and some representative visualization results are shown in section~\ref{sec_detail}, section~\ref{abla_study}, and section~\ref{sec_visualization}, respectively.

\section{Implementation Details}
\label{sec_detail}

In this section, we illustrate the detailed architectures of the feature enhancement module in the proposed CFFE and the CFNet framework with the backbone of the CPGNet~\cite{li2022cpgnet}.

\begin{figure}[!t]
	\centering
	\includegraphics[width=0.9\linewidth]{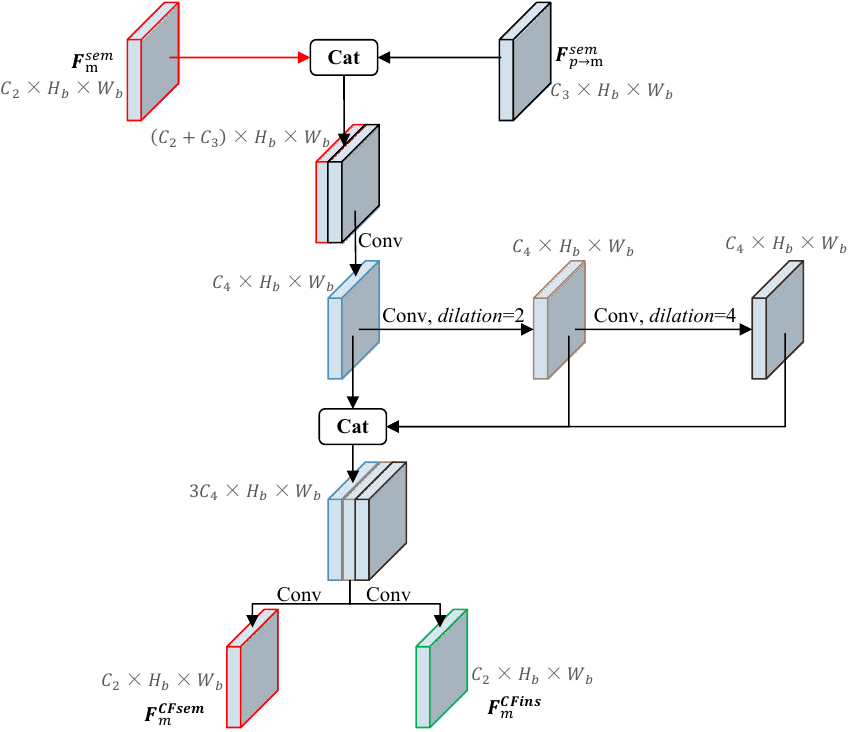}
	\caption{The feature enhancement module in the proposed CFFE. ``Conv'' represents a 2D convolution with $3 \times 3$ kernels, a batch normalization, and a ReLU layer. $dilation$ denotes the dilation coefficient of the 2D convolution and is set as 1 unless specified.}
	\label{fig_fem}
\end{figure}

\textbf{Feature enhancement module.}\quad As shown in Fig.~\ref{fig_fem}, the feature enhancement module (FEM) fuses the semantic feature maps $\boldsymbol{F}_m^{sem}$ and the re-projected shifted center feature maps $\boldsymbol{F}_{p \rightarrow m}^{sem}$ to generate center-focusing semantic feature maps $\boldsymbol{F}_m^{CFsem}$ and instance feature maps $\boldsymbol{F}_m^{CFins}$ ($m$ is the specific 2D view, such as RV~\cite{cortinhal2020salsanext, milioto2019rangenetplus, xu2020squeezesegv3}, BEV~\cite{lang2019pointpillars}, and polar view~\cite{zhang2020polarnet}.), which are used by the subsequent semantic and instance branches for more accurate predictions. Specifically, it first concatenates the two feature maps. Then, the concatenated feature maps undergo three convolution layers, where the dilation coefficients are set as 1, 2, and 4, respectively, to enlarge the receptive field. In the experiments, it is found that a larger receptive field can improve performance. Finally, the outputs of the three convolution layers are concatenated and then undergo two extra convolution layers to get semantic and instance feature maps, respectively. In our implementation, $C_2$ denotes the number of output channels of the corresponding 2D projection-based backbone. $C_3$ and $C_4$ are set as 64 and 48, respectively.

\begin{figure*}[!t]
	\centering
	\includegraphics[width=0.98\linewidth]{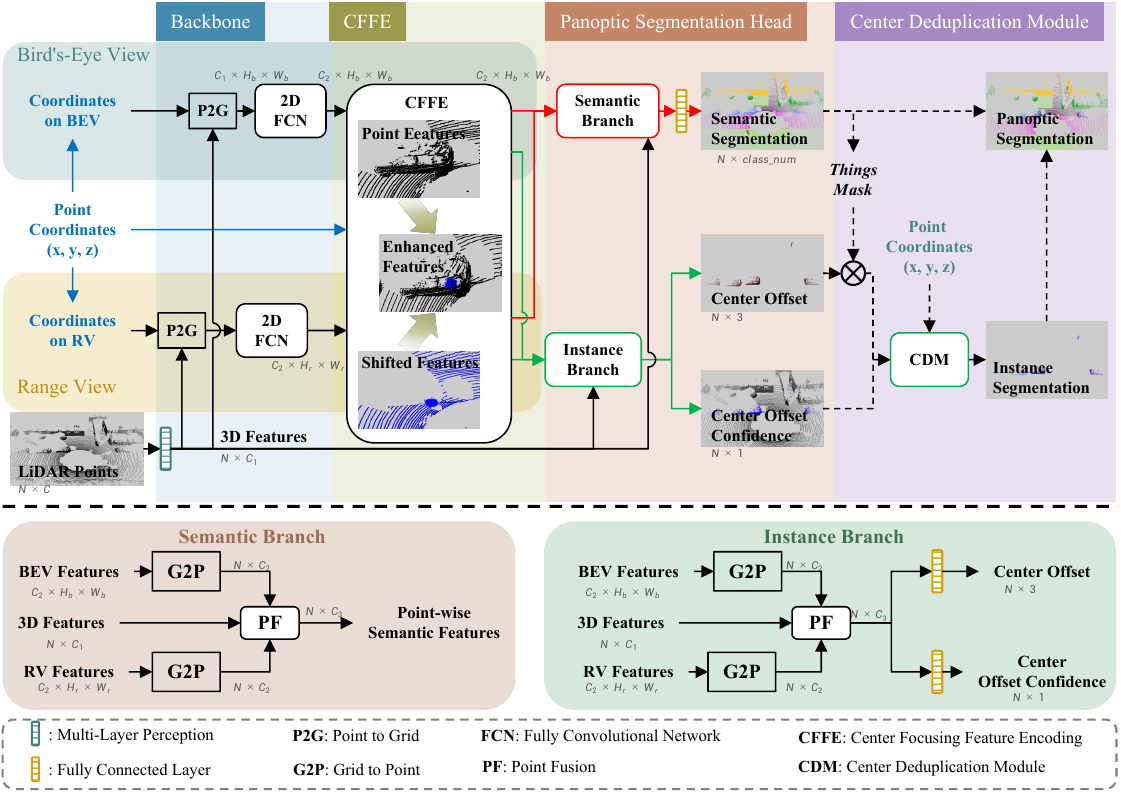}
	\caption{The overview of our CFNet with the backbone of the CPGNet~\cite{li2022cpgnet}.}
	\label{fig_cfnet_cpgnet}
\end{figure*}

\begin{figure*}[!t]
	\centering
	\includegraphics[width=0.98\linewidth]{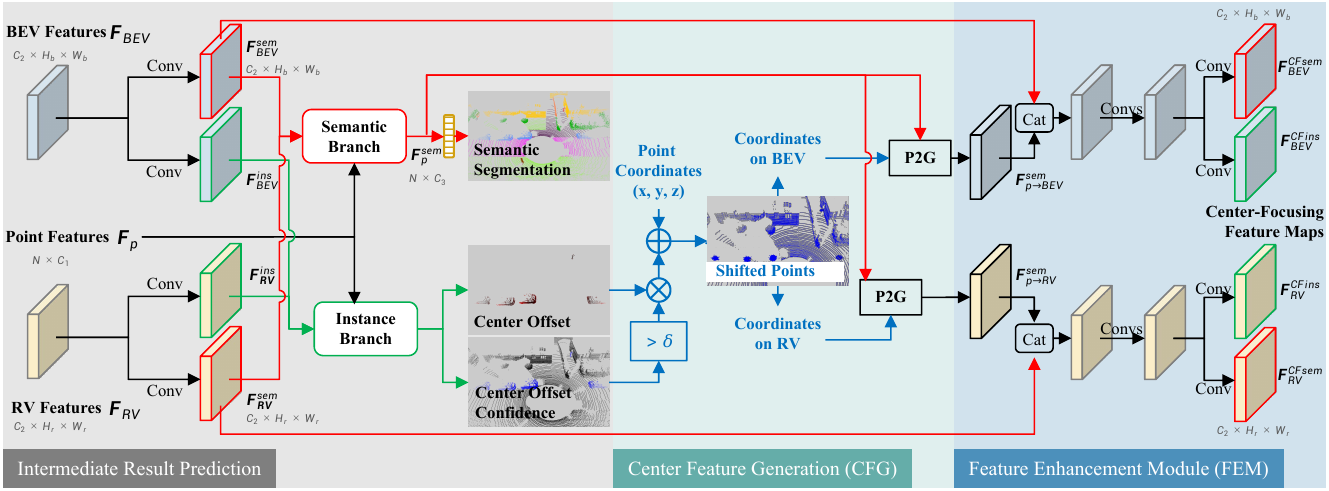}
	\caption{The proposed center focusing feature encoding (CFFE) that is integrated with the CPGNet~\cite{li2022cpgnet} backbone. The ``Conv'' represents a 2D convolution with $3 \times 3$ kernels, a batch normalization, and a ReLU layer. The details of the semantic branch and instance branch are shown in Fig.~\ref{fig_cfnet_cpgnet}. The blue arrows are coordinate-related operations.}
	\label{fig_cffe_cpgnet}
\end{figure*}

\textbf{CFNet with the backbone of the CPGNet.}\quad Fig.~\ref{fig_cfnet_cpgnet} presents the proposed CFNet with the backbone of the CPGNet~\cite{li2022cpgnet}, which is a powerful and efficient multi-view fusion backbone and consists of the 2D projection-based bird’s-eye view (BEV) and range view (RV) branches. Fig.~\ref{fig_cffe_cpgnet} shows the corresponding center focusing feature encoding (CFFE) that is integrated with the CPGNet~\cite{li2022cpgnet} backbone. These two modules are similar to those of the single-view backbone but add another view to alleviate the information loss during 2D projection.

As shown in Fig.~\ref{fig_cfnet_cpgnet}, for efficiency, only one stage version of the CPGNet is adopted in our CFNet. In the CPGNet, the P2G operation aims to project the LiDAR point features onto the BEV and RV feature maps. Specifically, $C_1$ and $C_2$ are set as 64. $H_b$ and $W_b$ are set as 600. $H_r$ and $W_r$ are set as 64 and 2048, respectively. The 2D FCN extracts features on each view. On the contrary to the P2G, the G2P operation transmits the features from each view back to the LiDAR points. The PF is responsible for fusing the features from the 3D points, BEV, and RV to generate point-wise features for the following predictions. For the details of the CPGNet backbone, please refer to the CPGNet~\cite{li2022cpgnet}.

\begin{table*}[!ht]
	\centering
	\resizebox{0.7\linewidth}{!}{
		\begin{tabular}{c|l|c|ccccc}
			
			\hline
			
			& Methods & Backbone & PQ & PQ$^\dag$ & PQ$^{Th}$ & PQ$^{St}$ & mIoU\\

			\hline \hline
			
			a & CFNet [Ours] & \multirow{3}{*}{CPGNet~\cite{li2022cpgnet}} & 62.7 & 67.5 & 70.0 & 57.3 & 67.4\\
			
			b & CFNet [Ours]; $dilation=1$ & & 62.2 & 66.7 & 68.5 & 57.2 & 67.1\\
			
			c & CFNet [Ours]; GT Offsets & & {\bf 65.5} & {\bf 69.7} & {\bf 76.4} & {\bf 57.5} & {\bf 69.5}\\
			
			\hline
		\end{tabular}
	}
	\caption{Ablation studies on the SemanticKITTI validation set. $dilation=1$ denotes that all dilation coefficients in the feature enhancement module (FEM) are set as 1. ``GT Offsets'' means that the center feature generation (CFG) generates the re-projected feature maps $\boldsymbol{F}_{p \rightarrow m}^{sem}$ according to the ground-truth center offsets instead of the predicted ones.}
	\label{table_ab_semkitti}
\end{table*}

\begin{table*}[!ht]
	\centering
	\resizebox{0.9\linewidth}{!}{
		\begin{tabular}{l|c|c|c|c|c|c|c|c|c|c|c|c|c}
			
			\hline
			distance threshold $d$ & 0 & 0.2 & 0.4 & 0.6 & 0.8 & 1.0 & 1.2 & 1.4 & 1.6 & 1.8 & 2.0 & 2.2 & 2.4\\
			\hline
			
			car & 20.4 & 64.3 & 90.4 & 94.2 & 94.9 & 95.1 & {\bf 95.2} & {\bf 95.2} & {\bf 95.2} & {\bf 95.2} & {\bf 95.2} & 95.0 & 94.6\\
			\hline
			
			truck & 4.7 & 29.2 & 40.6 & 59.5 & 71.3 & 71.7 & {\bf 71.9} & {\bf 71.9} & {\bf 71.9} & 71.5 & 70.8 & 70.0 & 68.9\\
			\hline
			
			person & 16.6 & 76.9 & 84.7 & {\bf 85.5} & {\bf 85.5} & 84.9 & 83.3 & 82.7 & 81.9 & 81.0 & 79.9 & 77.8 & 76.8\\
			\hline
			
			bicycle & 10.2 & 42.3 & 55.3 & 58.0 & {\bf 58.6} & 58.2 & 58.2 & 58.1 & 57.8 & 57.2 & 56.4 & 56.1 & 55.9\\
			\hline
			
		\end{tabular}
	}
	\caption{Different values of distance threshold $d$ in the proposed center deduplication module (CDM) on the SemanticKITTI validation set.}
	\label{table_distance_threshold}
\end{table*}

\begin{figure*}[!h]
	\centering
	\includegraphics[width=\linewidth]{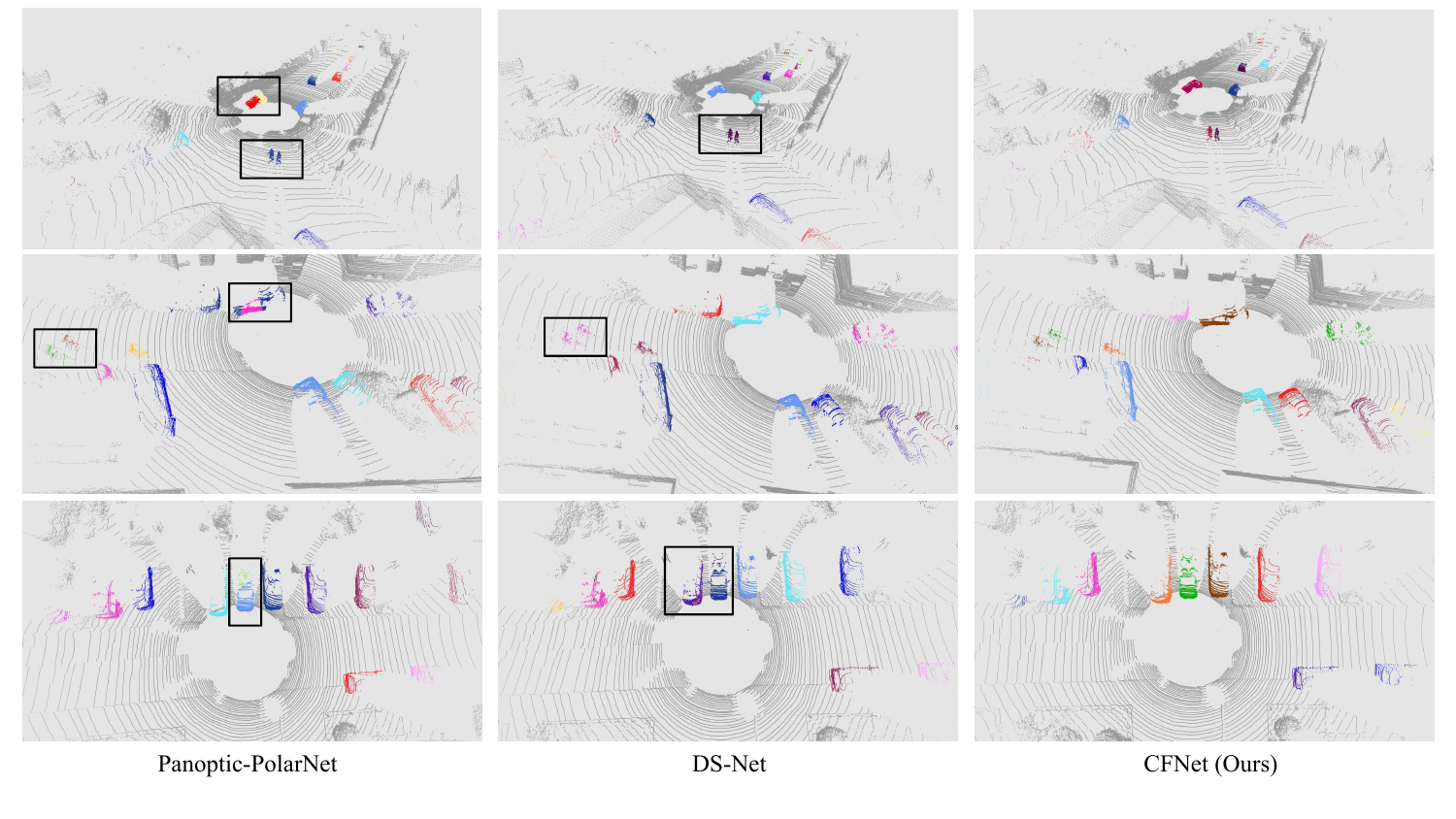}
	\caption{Comparison visualization results of the instance segmentation from the Panoptic-PolarNet~\cite{zhou2021panoptic}, DS-Net~\cite{hong2021lidar}, and our CFNet on the SemanticKITTI test set. The black box marks the region of interest.}
	\label{fig_comp_ins_seg}
\end{figure*}

\begin{figure*}[!ht]
	\centering
	\includegraphics[width=0.75\linewidth]{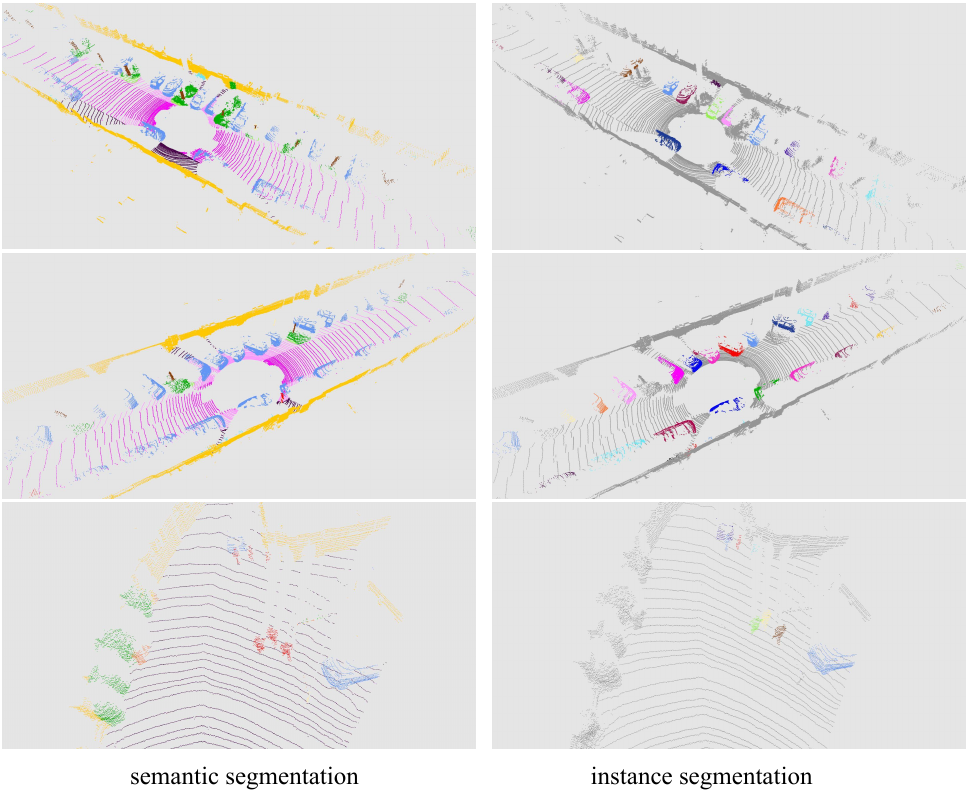}
	\caption{Visualization results of our CFNet on the SemanticKITTI test set. For semantic and instance segmentation, different colors represent different classes and instances, respectively.}
	\label{fig:cfnet_vis}
\end{figure*}

\section{Ablation studies}
\label{abla_study}

In this section, the ablative experiments are enriched for a comprehensive understanding of our CFNet.

\textbf{Different configurations of the CFFE.}\quad As shown in Table~\ref{table_ab_semkitti}, row (a) is the proposed CFNet with the backbone of the CPGNet~\cite{li2022cpgnet}. Row (b) is that all dilation coefficients in the feature enhancement module (FEM) are set as 1. Row (c) denotes that the center feature generation (CFG) generates the re-projected feature maps $\boldsymbol{F}_{p \rightarrow m}^{sem}$ according to the ground-truth center offsets instead of the predicted ones. It can be discovered that: 1) the larger receptive field results in the better performance (a,b); 2) the ground-truth center offsets facilitate the biggest performance improvements (a,c), which illustrates the importance and upper bound of the CFFE.

\begin{table}[!b]
	\centering
	\resizebox{\linewidth}{!}{
		\begin{tabular}{l|cccc|c}
			
			\hline
			
			Methods & PQ & PQ$^\dag$ & SQ & RQ & mIoU \\
			\hline \hline
			
			EfficientLPS~\cite{sirohi2021efficientlps} & 62.4 & 66.0 & 83.7 & 74.1 & 66.7\\
			
			Panoptic-PolarNet~\cite{zhou2021panoptic} & 63.6 & 67.1 & 84.3 & 75.1 & 67.0\\
			
			Panoptic-PHNet~\cite{li2022panoptic} & {\bf 80.1} & {\bf 82.8} & {\bf 91.1} & {\bf 87.6} & 80.2\\
			
			\hline
			{\bf CFNet~[Ours] w/CPGNet~\cite{li2022cpgnet}} & 79.4 & 81.6 & 90.7 & 87.0 & {\bf 83.6}\\
			\hline
		\end{tabular}
	}
	\caption{Comparison results on the nuScenes test set.}
	\label{table_nusc}
\end{table}

\textbf{Distance threshold $d$ on different classes.}\quad To figure out the effects of the distance threshold $d$, Table~\ref{table_distance_threshold} shows the distance threshold $d$ versus the PQ metric on some representative classes. The \emph{car} and \emph{truck} denote large objects, while the \emph{person} and \emph{bicycle} are small objects. The \emph{car} and \emph{truck} get the best PQ when the distance threshold $d$ is in the range of 1.2 to 1.6. The \emph{person} and \emph{bicycle} get the highest PQ when the distance threshold $d$ is set as 0.8. Thus, the optimal distance threshold $d$ varies from different classes. However, when $d$ is set as 0.8 in the main body, the performances of different classes are comparable to the optimal ones.

\textbf{Comparison results on the nuScenes test set.}\quad As shown in Table~\ref{table_nusc}, the proposed CFNet can be comparable with the state-of-the-art Panoptic-PHNet~\cite{li2022panoptic} on the nuScenes test set. However, the proposed CFNet runs much faster, as referred to the Table 3 of the main body.

\section{Visualization}
\label{sec_visualization}

In this section, our CFNet with the backbone of the CPGNet~\cite{li2022cpgnet} is inferred on the SemanticKITTI test set.

\textbf{Comparison visualization results.}\quad We run the official code of the Panoptic-PolarNet~\cite{zhou2021panoptic} and DS-Net~\cite{hong2021lidar} with the provided model parameters on the SemanticKITTI test set. For better visualization comparison, it only presents the instance segmentation results in Fig.~\ref{fig_comp_ins_seg}. It can be observed that the over-segmented and under-segmented problems frequently occur in the Panoptic-PolarNet and DS-Net, while our CFNet can avoid this problem. By the way, the over-segmented problem means that an instance is split into several parts and the under-segmented problem means that adjacent instances are predicted as a single instance.

\textbf{Visualization results.}\quad We present more visualization results of our CFNet on the SemanticKITTI test set in Fig.~\ref{fig:cfnet_vis}. Our CFNet can distinguish adjacent objects. Besides, the boundaries of instances can be accurately segmented.

{\small
	\bibliographystyle{ieee_fullname}
	\bibliography{egbib}
}
\end{document}